\documentclass[10pt,conference]{IEEEtran}
\IEEEoverridecommandlockouts

\usepackage{cite}
\usepackage{amsmath,amssymb,amsfonts}
\usepackage{algorithmic}
\usepackage{graphicx}
\usepackage{textcomp}
\usepackage{xcolor}
\usepackage{amsthm}
\usepackage{algorithm}
\usepackage{array}
\usepackage[caption=false,font=normalsize,labelfont=sf,textfont=sf]{subfig}
\usepackage{textcomp}
\usepackage{stfloats}
\usepackage{url}
\usepackage{verbatim}
\usepackage{cite}
\usepackage{mathabx}
\usepackage{comment}
\usepackage{booktabs}
\usepackage{multirow}
\def\BibTeX{{\rm B\kern-.05em{\sc i\kern-.025em b}\kern-.08em
    T\kern-.1667em\lower.7ex\hbox{E}\kern-.125emX}}

\IEEEoverridecommandlockouts
\IEEEpubid{\makebox[\columnwidth]{979-8-3315-7729-2/26/\$31.00~\copyright2026 IEEE \hfill} \hspace{\columnsep}\makebox[\columnwidth]{ }}

\begin{document}

\title{Hierarchical Asynchronous Federated Learning over Space-Air-Ground Integrated Networks
}

\author{Kai~Wang and Chee~Wei~Tan\\
College of Computing and Data Science, Nanyang Technological University, Singapore 639798\\
Email: wang1899@e.ntu.edu.sg, cheewei.tan@ntu.edu.sg
}

\maketitle
\IEEEpubidadjcol

\begin{abstract}
With the emerging 6G paradigm where AI services operate over decentralized and resource-constrained edge infrastructure, performance of distributed learning algorithms remains a significant challenge. This paper proposes HAFL, a Low Earth Orbit (LEO) satellite-assisted hierarchical asynchronous Federated Learning (FL) framework, enabling resource-limited terrestrial clients to collaborate via Ground-Satellite Links (GSLs). Within this framework, LEO satellites serve multiple roles: as edge servers aggregating and broadcasting models, as clients accelerating training, and as high-throughput relays managing inter-satellite routing through Inter-Satellite Links (ISLs). Unlike conventional synchronous Jacobian-type updates, HAFL employs asynchronous Gauss-Seidel-type updates to enhance convergence speed. We provide theoretical convergences for convex, strongly convex, and nonconvex objectives, respectively, and formulate optimization problems to jointly determine optimal hyperparameters to minimize global training latency. Evaluations demonstrate that HAFL achieves faster initial convergence and comparable final performance for benchmark DNNs and datasets, offering promising FL systems across remote areas with 6G technologies, but much less global rounds and communication cost.
\end{abstract}

\section{Introduction}\label{sec:sec1}

\IEEEPARstart{T}{he} sixth-generation (6G) wireless technologies are envisioned to push beyond the capabilities of 5G by offering ultra-high data rates, extremely low latency, pervasive Artificial Intelligence (AI), and native support for ubiquitous connectivity across terrestrial, aerial, and satellite platforms. In particular, the increasing demand for AI-driven applications at the network edge has motivated distributed learning paradigms capable of operating in constrained and heterogeneous environments. In the 6G landscape, where the network edge plays an important role as data sources and model trainers, designing distributed algorithms that are both communication-efficient and convergence-guaranteed presents a unique challenge.

Federated Learning (FL) \cite{fl1}, a distributed machine learning framework where edge devices collaboratively train a global model without sharing raw data, has emerged as a promising technique and has received increasing attention. By leveraging the computation capabilities of distributed equipment \cite{fl2}, the emerging 6G presents a promising enabler for FL over remote areas lacking terrestrial infrastructure and served only by satellites. In particular, we consider the situation where clusters of clients are served by Low Earth Orbit (LEO) satellites. It is beneficial to involve all clients since they have individual datasets that could be highly non-i.i.d. across different clusters; meanwhile, the clients might have disproportionate computing capabilities that suggests a potential split of training task into submodels accordingly. However, distributed clusters of clients require intensive communication, suffering from the scalability issue with single-server FL. To mitigate this, Hierarchical FL (HFL) \cite{hfl1,hfl2,hfl3,hfl5,hfl7,hfl9,hfl10,hfl11,hfl12} was proposed that introduces edge servers as the middle level aggregators in the hierarchy and reduces direct communication with the cloud server.

Over the past decade, satellite link capacity (e.g., Starlink \cite{starlink}) has outpaced edge training performance (e.g., iPhone \cite{iphone}), in that satellite bandwidth capacity has scaled extremely fast, whereas edge device compute has improved more modestly (especially for training). At the same time, Gauss-Seidel-type updates empirically converge faster than Jacobian-type updates but require inter-cluster communication over distant nodes. Motivated by these observations, we leverage the ultra-high transmission rate of LEO satellites in 6G and propose an asynchronous HFL-based architecture that partitions the model into disjoint submodels whose sizes are optimized according to clients' computing and transmission capabilities, offloads insensitive data to satellite to leverage the computing capabilities of satellites, and use asynchronous (i.e., Gauss-Seidel-type) submodel updates for faster and guaranteed convergence. Our main contributions are summarized as follows.
\begin{itemize}
    \item We propose Hierarchical Asynchronous Federated Learning (HAFL) to enhance the training efficiency of FL with Gauss-Seidel-type submodel updates, which sequentially updates submodels and accelerates convergence behavoir.
    \item We leverage LEO satellites in multiple orbit planes to engage remotely distributed clients with local datasets. The LEO satellites are versatile to perform submodel training with offloaded data like an edge client, orchestrate intra-cluster submodel aggregation via Ground-Satellite Links (GSLs), and transmit submodels to remote LEO satellites via Inter-Satellite Links (ISLs).
    \item We formulate a network latency minimization problem using iterative Block Coordinate Descent (BCD), converting it into multiple LPs to obtain system parameters.
    \item We prove the convergence under various convexity assumptions, showing that HAFL admits a convergence rate of $\mathcal{O}(1/\sqrt{t})$ and $\mathcal{O}(1/t)$ with $t$ global updates for convex and strongly convex objectives, and ensures a vanishing gradient norm for nonconvex objectives.
    \item We evaluate HAFL with benchmark DNN architectures and datasets to demonstrate the performance over baselines in terms of training and communication costs.
\end{itemize}

The remainder of the paper is organized as follows. Section \ref{sec:sec2} presents LEO satellite models. Section \ref{sec:sec3} describes the communication aspect and solves LP-based optimization. Section \ref{sec:sec4} describes the training aspect and convergence results. Section \ref{sec:sec5} presents experiment results and analysis. Section \ref{sec:sec6} concludes with discussion and future work.

\section{LEO Satellite Models}\label{sec:sec2}

\subsection{The Satellite Constellation}

The hierarchical learning and communication system consists of cloud servers $\mathcal{K}$, edge servers $\mathcal{S}$, client devices $\mathcal{J}$, and clusters $\{\mathcal{C}_i\}_{i\in\mathcal{I}}$, where each client $j \in \mathcal{J}$ is associated with exactly one cluster $\mathcal{C}_i$, such that $\mathcal{J} = \bigcup_i \mathcal{C}_i$ and $\mathcal{C}_i \bigcap \mathcal{C}_{i'} = \varnothing$ for all $i \neq i'$, and each cluster $\mathcal{C}_i$ is associated with an edge server $S_i \in \mathcal{S}$, which may concurrently serve multiple clusters.

We adopt the LEO megaconstellation model with a set $\mathcal{N}$ of orbit planes \cite{satellite}, where each orbit $n\in\mathcal{N}$ consists of $O_n$ co-orbital LEO satellites around the Earth in sequence. Since LEO satellites are not geostationary, related work \cite{satellite2, satellite3} proposed to model the LEO satellite megaconstellation as a fixed topology snapshot within time slot $T$, and the satellite currently serving the cluster must hand over to the next co-orbital satellite to ensure continuity of service. Given altitude $h_n$ of orbit $n$ and Earth radius $R_E$, the distance from orbit $n$ to Earth center is $h_n^*=h_n+R_E$, and the position satellite $i$ in orbit $n$ in geocentric coordinates is given by \cite{satellite2}
\begin{align*}
    x_i=&\ h_n^*\cos\alpha_i\cos\gamma_n-h_n^*\sin\alpha_i\cos\kappa_n\sin\gamma_n,\\
    y_i=&\ h_n^*\cos\alpha_i\sin\gamma_n+h_n^*\sin\alpha_i\cos\kappa_n\sin\gamma_n,\\
    z_i=&\ h_n^*\sin\alpha_i\cos\kappa_n,
\end{align*}
where $\alpha_i$, $\gamma_n$, and $\kappa_n$ denote the right ascension of ascending node of satellite $i$ in orbit $n$, the longitude angle, and the orbital inclination of orbit $n$, respectively. The distance $d_{i,i'}$ between satellites $i$ and $i'$ is given by
\begin{align*}
    d_{i,i'}=\sqrt{(x_i-x_{i'})^2+(y_i-y_{i'})^2+(z_i-z_{i'})^2},
\end{align*}
and the communication radius of satellite $i$ in orbit $n$ is
\begin{align*}
    d^\text{comm}_i=2\sqrt{(h_n^*)^2-R_E^2}.
\end{align*}
Inter-orbit ISLs exist only when satellites $i$ and $i'$ are both within each other's communication radius, i.e., $d_{i,i'}\leq\min(d^\text{comm}_i,d^\text{comm}_{i'})$, and intra-orbit ISLs are only available between adjacent co-orbit satellites \cite{satellite2}. With these available ISLs, we can construct the communication graph.

\subsection{The Satellite Handover and Communication Model}

LEO satellites are non-geostationary that provide coverage to a cluster only for coverage time $T$, after which the connection and task is handed over to a neighboring satellite in time $\tau^{\text{trans}}_{S,i}$. We adopt the satellite handover model \cite{gts}
\begin{align}\label{eq:handover}
    T^\text{eff}_{S,i}=T^\text{raw}_{S,i}+(N_i+1)\cdot\tau^\text{trans}_{S,i},\ N_i=\big\lfloor\frac{T^\text{raw}_{S,i}}{T-\tau^\text{trans}_{S,i}}\big\rfloor,
\end{align}
where $T^\text{raw}_{S,i}$ is the raw task time, and $T^\text{eff}_{S,i}$ is the actual task time with handover. The rationale is that satellite $i$ must hand over by reserving time $\tau^\text{trans}_{S,i}$ at the end of slot $T$ to transmit its model and dataset to an adjacent satellite. Satellite $i$ can communicate to satellite $j$ via ISL with rate given by
\begin{align}\label{eq:srate}
    R^\text{ISL}_{i,j}=B_{S,i}\log_2\big(1+\frac{P_{S,i}|h_{i,j}|^2}{B_{S,i}N_0}\big),
\end{align}
where $B_{S,i}$ is the bandwidth for satellite $i$, $N_0$ is the white noise power spectral density, $P_{S,i}$ is satellite transmit power, and $h_{i,j}$ is the corresponding channel coefficient. For ISLs where free-space LoS channels dominate, $h_{i,j} = \sqrt{\beta_{i,j}}$ and
\begin{align}\label{eq:lossmodel}
    \beta_{i,j}=\beta_0-10\cdot\alpha^{\text{ISL}}\cdot\log_{10}\big(\frac{d_{i,j}}{d_0}\big),
\end{align}
where $\beta_0 = -30$dB denotes the large-scale pathloss coefficient at a reference distance of $d_0 = 1$m, $d_{i,j}$ denotes the distance between satellites $i$ and $j$, and $\alpha^{\text{ISL}}\!=\!2.00$ denotes the pathloss exponent for free-space ISLs.

We model GSL between satellite and client in urban areas with large-scale and small-scale fading effects given by $h_{i,j}=\sqrt{\beta_{i,j}}u_{i,j}$, where $\beta_{i,j}$ is the large-scale pathloss coefficient and $u_{i,j}\sim\mathcal{CN}(0,1)$ captures Rayleigh fading. For urban areas, we replace $\alpha^{\text{ISL}}$ in Eq. \eqref{eq:lossmodel} with $\alpha^{\text{GSL}}=3.50$. The energy consumption for transmission via ISL is
\begin{align}\label{eq:senergy}
    E^\text{ISL}_{i,j} = P_{S,i} \cdot \tau^\text{ISL}_{i,j}.
\end{align}
Formulas similar to Eqs. \eqref{eq:srate}-\eqref{eq:senergy} hold for GSL channels.

\section{Communication Aspect and Optimization}\label{sec:sec3}

In this section, we describe each stage of the communication system that occurs within a single global round.

\subsection{Preprocessing: Data and Model Offloading to Satellites}

Before global round $t=0$, each client $j\!\in\!\mathcal{C}_i$ offloads a subset of non-sensitive data $\hat{\mathcal{D}}_j^s\subset\mathcal{D}_j^s$ to satellite $i$ (i.e., edge server $i$). Define $\alpha_j$ as the portion of offloaded data and define $\alpha_j^{\max}=|\hat{\mathcal{D}}_j^s|/|\mathcal{D}_j|$ as the maximum portion of offloaded data. The cloud server allocates submodel $S_i$ to cluster $i$ and all clients $j\in\mathcal{C}_i$ are synchronized with the same submodel $S_i$. These steps occur at all nodes once before global round $t=0$.

\subsection{Cluster Round: Satellite Side Update}

We decide mini-batch sizes that scale proportionally to the local dataset size, enabling effective utilization of all on-device data under a fixed number of training rounds across datasets of varying sizes. For simplicity, we express the raw task time for $H$ rounds of submodel updates at satellite $i$ as
\begin{align}\label{eq:T_clusterraw_Si}
    T_{S,i}^\text{cluster,raw}=\frac{Hm_{S,i}\gamma_{S,i}\sum_{j\in\mathcal{C}_i}\alpha_j|\mathcal{D}_j|}{f_{S,i}},
\end{align}
where $f_{S,i}$ is the CPU frequency, $m_{S,i}$ is the number of CPU cycles per sample, $\gamma_{S,i}$ is the portion of samples a mini-batch contains. The handover time is
\begin{align}\label{eq:tau_trans_Si}
    \tau^{\text{trans}}_{S,i}=\frac{|S|+q\cdot\sum_{j\in\mathcal{C}_i}\alpha_j|\mathcal{D}_j|}{R^{\text{ISL}}},
\end{align}
where $|S|$ is the size of the entire model since $S$ is always needed to train any submodel $S_i$, $q$ is the size of each data sample, and $R^{\text{ISL}}$ is the ultra-high ISL rate between adjacent satellites and is assumed to be constant. The satellite handover gives the satellite side latency as
\begin{align}\label{eq:T_cluster_Si}
    T_{S,i}^\text{cluster}=T_{S,i}^\text{cluster,raw}+\big(\big\lfloor \frac{T_{S,i}^\text{cluster,raw}}{T-\tau^\text{trans}_{S,i}}\big\rfloor+1\big)\cdot{\tau^\text{trans}_{S,i}}.
\end{align}
The energy consumption per cluster round at satellite $i$ is the sum of the energy used for CPU and transmission
\begin{align}\label{eq:E_cluster_Si}
    E^\text{cluster}_{S,i}=\kappa T_{S,i}^\text{cluster,raw} f_{S,i}^3+P_{S,i}\cdot \big(\big\lfloor\frac{T_{S,i}^\text{cluster,raw}}{T-\tau^\text{trans}_{S,i}}\big\rfloor+1\big)\cdot{\tau^\text{trans}_{S,i}},
\end{align}
where the RHS terms denote CPU computation and transmission energy consumptions, respectively. The energy constraint requires that the battery level of satellite $i$ should always remain above a given threshold. That is,
\begin{align*}
    \phi^{\text{initial}}_{S,i}-E^{\text{cluster}}_{S,i}+{P^{\text{sun}}_{S,i}}\cdot{T^{\text{cluster}}_{S,i}}\geq\phi^{\min}_{S,i},
\end{align*}
where $P^{\text{sun}}_{S,i}$ means the solar energy recharge rate, and $\phi^{\text{initial}}_{S,i}$ and $\phi^{\min}_{S,i}$ denote the initial battery level charged and the minimum battery level to function, respectively. Since monitoring battery level is difficult, we impose a more stringent constraint that the energy consumption should not exceed the energy recharge from the sun during a cluster round. That is,
\begin{align}\label{eq:E_cluster_Si_constraint}
    E^{\text{cluster}}_{S,i}\leq{P^{\text{sun}}_{S,i}}\cdot{T^{\text{cluster}}_{S,i}},
\end{align}
which corresponds to setting $\phi^{\text{initial}}_{S,i}=\phi^{\min}_{S,i}$.

\subsection{Cluster Round: Client Side Update and Upload}

The time $T_{C,j}^\text{cluster}$ to compute $H$ local updates at client $j$ is
\begin{align}\label{eq:T_cluster_Cj}
    T_{C,j}^\text{cluster}=\frac{Hm_{C,j}\gamma_{C,j}(1-\alpha_j)|\mathcal{D}_j|}{f_{C,j}},
\end{align}
where $f_{C,j}$ is the CPU frequency of client $j$, $m_{C,j}$ is the number of CPU cycles to process one sample, and $\gamma_{C,j}$ is the portion of samples a mini-batch contains. The corresponding energy consumption is given by
\begin{align}\label{eq:E_cluster_Cj}
    E_{C,j}^\text{cluster}=\kappa{H}m_{C,j}\gamma_{C,j}(1-\alpha_j)|\mathcal{D}_j|f_{C,j}^2.
\end{align}
The communication time for client $j$ to upload the submodel to satellite $i$ is
\begin{align}\label{eq:tau_agg_Cj}
    \tau_{C,j}^\text{agg}=\frac{|S_i|}{R^{CS}_{j,i}},\ {R^{CS}_{j,i}}=B_j\log_2\big(1+\frac{P_{C,j}|h_{j,i}|^2}{B_jN_0}\big).
\end{align}
The corresponding energy consumption for submodel transmission at client $j$ becomes
\begin{align}\label{eq:E_agg_Cj}
    E^\text{agg}_{C,j}=P_{C,j}\cdot\tau^\text{agg}_{C,j}.
\end{align}
\begin{figure}[t]
  \centering
  \includegraphics[width=0.50\textwidth]{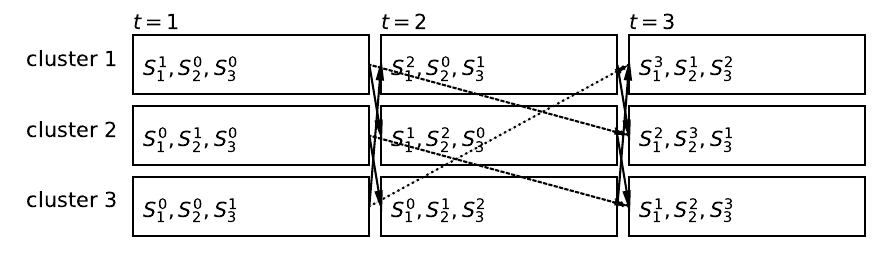}
  \caption{An illustration of the submodel transmission scheme. In this scheme, solid lines denote transmission in Stage Two, and dotted lines denote transmission in Stage Three. For example, cluster 2 needs $S^1_1$ to train $S^2_2$ and hence $S^1_1$ must be transmitted before $t=2$, corresponding to Stage Two. However, cluster 3 needs $S^1_1$ to train $S^3_3$, which is time-insensitive and can be arranged to transmit during training, corresponding to Stage Three.}
  \label{fig:transmissionscheme}
  \vspace{-20pt}
\end{figure}
\!\!The analysis of client-side updates and uploads alone is more nuanced, and we adopt the client latency model \cite{gts} to capture this complexity. The duration $T^{\text{total}}_{i}$ measures the time elapsed from the start of the cluster round to the point when satellite $i$ has aggregated submodels from all clients, as computed by Eq. \eqref{eq:T_total_i} under three cases in Eq. \eqref{eq:T_total_cases}. The rationale behind Eqs. \eqref{eq:T_total_i} and \eqref{eq:T_total_cases} is explained in literature \cite{gts}.

\begin{figure*}[!t]
\vspace{5pt}
\begin{align}\label{eq:T_total_i}
T^\text{total}_i = 
\begin{cases}
T \cdot N_i + \max_{j \in \mathcal{C}_i} \big\{ \tau^{\text{agg}}_{C,j} \big\}, & \text{ (case $a$)$\qquad\qquad\qquad\qquad\qquad\qquad\qquad$} \\
\max_{j \in C_i} \big\{ \max \big( T \cdot \big\lfloor \dfrac{T^{\text{cluster}}_{C,j'}}{T} \big\rfloor, T^{\text{cluster}}_{C,j} \big) + \tau^{\text{agg}}_{C,j} \big\}, & \text{ (case $b$)$\qquad\qquad\qquad\qquad\qquad\qquad\qquad$} \\
T \cdot \big( \big\lfloor \dfrac{T^{\text{cluster}}_{C,j'}}{T} \big\rfloor + 1 \big) + \max_{j \in C_i} \big\{ \tau^{\text{agg}}_{C,j} \big\}, & \text{ (case $c$)$\qquad\qquad\qquad\qquad\qquad\qquad\qquad$}
\end{cases}
\qquad\\
\begin{cases}\label{eq:T_total_cases}
    \text{case $a$:} & T^{\text{cluster}}_{C,j'}\leq T \cdot N_i \\
    \text{case $b$:} & T^{\text{cluster}}_{C,j'}> T \cdot N_i\text{ and }\max_{j \in C_i} \big\{ \max \big( T \cdot \big\lfloor \dfrac{T^{\text{cluster}}_{C,j'}}{T} \big\rfloor, T^{\text{cluster}}_{C,j} \big)\! +\! \tau^{\text{agg}}_{C,j} \big\}\!\leq\! T \cdot \big( \big\lfloor \dfrac{T^{\text{cluster}}_{C,j'}}{T} \big\rfloor\! +\! 1 \big) \\
    \text{case $c$:} & T^{\text{cluster}}_{C,j'}> T \cdot N_i\text{ and }\max_{j \in C_i} \big\{ \max \big( T \cdot \big\lfloor \dfrac{T^{\text{cluster}}_{C,j'}}{T} \big\rfloor, T^{\text{cluster}}_{C,j} \big)\! +\! \tau^{\text{agg}}_{C,j} \big\}\!>\! T \cdot \big( \big\lfloor \dfrac{T^{\text{cluster}}_{C,j'}}{T} \big\rfloor\! +\! 1 \big) 
\end{cases}
\end{align}
\vspace{-20pt}
\end{figure*}

The latency per cluster round $T^\text{cluster}_i$ is the larger of satellite latency and client latency, plus synchronization delay, that is,
\begin{align*}
    T^\text{cluster}_{i}=\max(T_{S,i}^\text{cluster},T^\text{total}_i)+\tau^\text{sync1}_i,
\end{align*}
where $\tau^\text{sync1}_i$ is the additional latency to broadcast aggregated submodel to all clients in cluster $i$. The cluster rounds are repeated $E$ times on both satellite and client sides, after each of which a consensus submodel is aggregated at the satellite and broadcast to all clients, requiring a total delay of $E\cdot T^\text{cluster}_{i}$.

\subsection{Global Round: Inter-Cluster Propagation}\label{sec:sec3d}

The LEO megaconstellation has $\mathcal{N}$ orbit planes and three GEO satellites \cite{satellite}. The submodel $S_i$ trained during global round $t$ is denoted as $S^t_i$. There are three communication stages in global round $t$: satellite $i$ broadcasts $S^{t-1}_{i-1}$ to clients in its cluster that also upload the local models to satellite $i$ (stage 1), satellite $i$ transmits $S^{t-1}_i$ to satellite $(i+1)$ (stage 2), and satellite $i$ transmits $S^{t-1}_i$ to all other satellites (stage 3). Note that stage 3 is arranged concurrently with training and is therefore not time-critical. An illustration of the transmission scheme of the submodel is shown in Fig. \ref{fig:transmissionscheme}.

\subsection{Iterative Global Latency Minimization}

The objective to minimize is the latency of a global round $T^{\text{glob}}$ is given by
\begin{align}\label{eq:glob}
    T^{\text{glob}}=E\cdot \textstyle{\max}_iT^{\text{cluster}}_i.
\end{align}
In minimizing $T^{\text{glob}}$, we use the assumption (e.g., in \cite{gts}) that sync delays $\tau^{\text{sync}1}_i$ are short whose optimization contributes marginally to the global latency minimization. The objective is an implicit function of $\boldsymbol{\alpha}$, $\mathbf{f}_S$, $\mathbf{B}_C$, and $\mathbf{S}$. The global latency minimization problem is equivalent to
\begin{align}
    \min_{\boldsymbol{\alpha}, \mathbf{f}_S, \mathbf{B}_C, \mathbf{S}}&\ T^{\text{glob}} \label{eq:Tglob} \\
    \text{s.t.:}&\ 0 \leq \alpha_j \leq \alpha_j^{\max},\ \forall j \in \mathcal{C}_i, \ \forall i \in \mathcal{I} \label{eq:Tglob1} \\
    &\ f_{S,i}^{\min} \leq f_{S,i} \leq f_{S,i}^{\max},\ \forall i \in \mathcal{I} \label{eq:Tglob2} \\
    &\ \textstyle\sum_{j \in \mathcal{C}_i} B_{C,j} \leq B_i,\ \forall j \in \mathcal{C}_i, \ \forall i \in \mathcal{I} \label{eq:Tglob3} \\
    &\ E_{C,j}^{\text{cluster}} + E_{C,j}^{\text{agg}} \leq \delta_j, \ \forall j \in \mathcal{C}_i,\ \forall i \in \mathcal{I} \label{eq:Tglob4} \\
    &\ E_{S,i}^{\text{cluster}} \leq P_i^{\text{sun}} \cdot T_{S,i}^{\text{cluster}}, \ \forall i \in \mathcal{I} \label{eq:Tglob5} \\
    &\ \textstyle\sum_{j \in \mathcal{C}_i} \alpha_j |\mathcal{D}_j| \leq A_i^{\max}, \ \forall i \in \mathcal{I}, \label{eq:Tglob6} \\
    &\ \textstyle\sum_{i\in\mathcal{I}}|S_i|=|S|. \label{eq:Tglob7}
\end{align}

We propose a Block Coordinate Descent (BCD) algorithm to address the nonconvex problem in Eqs. \eqref{eq:Tglob}-\eqref{eq:Tglob7}, which iteratively optimizes over $\boldsymbol{\alpha}=[\alpha_{j}]_{j\in\mathcal{C}_i,i\in\mathcal{I}}$, $\mathbf{f}_S=[f_{S,i}]_{i\in\mathcal{I}}$, $\mathbf{B}_C=[B_{C,j}]_{j\in\mathcal{C}_i,i\in\mathcal{I}}$ and $\mathbf{S}=[S_i]_{i\in\mathcal{I}}$, resp. Given $t$, we can check the feasibility of $T^{\text{cluster}}_i\leq t$ with all constraints in Eqs. \eqref{eq:Tglob1}-\eqref{eq:Tglob7} using LP-solvers. Then we use bisection search on $t$ to find the minimum feasible $t$ that corresponds to the solution. By subdividing the discussion into cases ($a$), ($b$), and ($c$) in Eqs. \eqref{eq:T_total_i}-\eqref{eq:T_total_cases}, to check the feasibility of $T^{\text{cluster}}_i\ \!{\leq}\ \!t$, we can sequentially check that of $T^{\text{cluster}}_{S,i}+\tau^{\text{sync}}_i\ \!{\leq}\ \!t$, $T^{\text{total}}_i+\tau^{\text{sync}}_i\ \!{\leq}\ \!t$ under cases ($a$), ($b$), and ($c$), resp., where the entire procedure results in a number of feasibility tests \cite{gts}. We have to omit the details due to space limits.

\section{Training Aspect and Convergence Results}\label{sec:sec4}

We denote the portion of model parameters $\mathbf{x}$ associated with block $i$ (i.e., submodel $S_i$) and client $j$ at iteration $(t,e,h)$ by $\mathbf{x}^{t,e,h}_{i,j}$, where $t\in[T], e\in[E], h\in[H], i\in\mathcal{I}, j\in\mathcal{C}_i$. We use the objective decomposition $\Phi(\mathbf{x})=F(\mathbf{x})+R(\mathbf{x})$ \cite{bsg}, where $F(\mathbf{x})=\mathbb{E}_{\xi}[f(\mathbf{x};\xi)]$ and $R(\mathbf{x})=\sum_{i\in\mathcal{I}}r_i(\mathbf{x}_i)$ are the empirical loss and the structural loss, respectively. We can express the essential steps for client updates, cluster updates, and global updates in HAFL as follows.

\subsection{Updates}

\noindent\textit{1) Client update}. The objective function for client $j\in\mathcal{C}_i$ is defined with local data not offloaded to edge server $i$ to be
\begin{align*}
    \Phi_{i,j}(\mathbf{x})=\mathbb{E}_{\xi\sim{\mathcal{D}_j}}[\Phi_i(\mathbf{x};\xi)],
\end{align*}
where the subscript $i$ in $\Phi_i$ and $\Phi_{i,j}$ denotes the gradient is only with respect to $\mathbf{x}_i$. Each client $j\in\mathcal{C}_i$ computes $H$ mini-batch stochastic gradients with respect to submodel parameters $\mathbf{x}^{t,e,h}_{i,j}$, and updates the local submodel for $H$ steps via SGDs,
\begin{align*}
    \mathbf{x}^{t,e,h+1}_{i,j}=\mathbf{x}^{t,e,h}_{i,j}-\alpha^t\cdot\nabla_{\mathbf{x}_i}\Phi_{i,j}(\mathbf{x}^{t,e,h}_{i,j}),\ \forall h\in[H].
\end{align*}

\noindent\textit{2) Cluster update}. At the end of the $H$-th local update, each client $j$ uploads submodel $\mathbf{x}^{t,e,H}_{i,j}$ to edge server $i$. Then edge server $i$ aggregates the client submodels with
\begin{align*}
    \mathbf{x}^{t,e+1}_{i}=\frac{1}{\sum_{j\in\mathcal{C}_i}|\mathcal{D}_j|}\textstyle\sum_{j\in\mathcal{C}_i}|\mathcal{D}_j|\mathbf{x}^{t,e,H}_{i,j},\ \forall e\in[E]
\end{align*}
and broadcasts the aggregated submodel back to clients for subsequent cluster round $e+1$.

\noindent\textit{3) Global update}. At the end of the $E$-th cluster update, edge server $i$ transmits the aggregate submodel to edge server $i+1$. This stands in sharp contrast with conventional hierarchical FL frameworks where edge server $i$ uploads the aggregated model to the cloud to update the global model.

\subsection{Convergence Results}

Our theoretical analysis for convergence focuses on convex, strongly convex, and smooth and nonconvex objectives, respectively, under generally non-i.i.d. dataset distributions. We adopt standard assumptions in the FL literature.

\noindent\textbf{Theorem 1} (general convex functions). Let $\{\mathbf{x}^t\}$ be generated from HAFL with $\alpha^t=\frac{\theta}{\sqrt{t}}<\frac{1}{L}$ for some constant $0<\theta<\frac{1}{L}$, where $L$ is the Lipschitz constant. If $F$ and $r_i$'s are convex, $\mathbf{x}^*$ is the optimal solution, then for some constant $D(\mathbf{x}^*)>0$,
\begin{align*}
\mathbb{E}\big[\Phi(\tilde{x}^T)-\Phi(x^*)\big]\leq \frac{\theta(\ln T+1)}{\sqrt{T+1}}D(x^*)+\frac{\|x^*-x^0\|^2}{2EH\theta\sqrt{T+1}},
\end{align*}
which coincides with the well-known optimal rate of convergence $\mathcal{O}(1/\sqrt{t})$ under general convex setting for SGD.

\noindent\textbf{Theorem 2} (strongly convex functions). Let $\{\mathbf{x}^t\}$ be generated from HAFL with $\alpha_t\! =\! \frac{\theta}{t} < \frac{1}{L}$ for some constant $\theta$, where $L$ is the Lipschitz constant. If $F$ and $r_i$'s are convex, $\mathbf{x}^*$ is the optimal solution, and $\Phi$ is strongly convex with modulus $\mu > 0$, then for some constant $D(\mathbf{x}^*)>0$ and for $T\geq(\lfloor A\rfloor+1)/(EH)$ with $A={EH\mu\theta}/{(2+2\mu\theta)}$, we have
\begin{align*}
    &\mathbb{E}[\|x^T-x^*\|^2]\leq\frac{c}{T},
\end{align*}
where
\begin{align*}
    &c= \begin{cases}4EHD(x^*)\theta^2/(A-1)\ (A>1), \\ \max \big\{2EHD(x^*)\theta^2 / A, \mathbb{E}[\|x^1-x^*\|^2]\big/(EH)\} \ (A \leq 1),\end{cases}
\end{align*}
\begin{align*}
\mathbb{E}[\|\mathbf{x}^t\!-\!\mathbf{x}^*\|^2]\!\leq\!\frac{1}{t}\max\{\frac{2DEH\theta(1+sEH\mu\theta)}{sEH\mu},\|\mathbf{x}^{0}\!-\!\mathbf{x}^*\|^2\},
\end{align*}
which has the same optimal rate of convergence $\mathcal{O}(1/t)$ under strongly convex setting as that of SGD.

\noindent\textbf{Theorem 3}. Let $\left\{\mathbf{x}^t\right\}$ be generated from HAFL with $r_i=0$ and $\mathcal{X}_i=\mathbb{R}^{n_i}$ and with positive scalars $\alpha^t$ that satisfies $\alpha^t \geq \alpha^{t+1}$,
$\textstyle\sum_{t=0}^{\infty}\alpha_t=+\infty$, and $\sum_{t=0}^{\infty}\alpha_t^2<+\infty$. Then
\begin{align*}
    \lim_{t\to\infty}\mathbb{E}[\|g^{t,e,h}_{i,j}\|]=0,\ \forall e, h,\ \forall i, j,
\end{align*}
which ensures vanishing gradient norm $\mathit{o}(1)$.

Rigorous proofs are omitted due to space limit and will be provided in an upcoming extended version of this work.

\begin{table*}[!t]
\centering
\caption{Simulation Setting}
\vspace{10pt}
\label{tab:simulationsetting}
\begin{tabular}{|c|c||c|c||c|c|}
\hline
\textbf{Parameter} & \textbf{Value} & \textbf{Parameter} & \textbf{Value} & \textbf{Parameter} & \textbf{Value} \\
\hline
$f_{S,i}$ & $1$ GHz & $m_{S,i}$ & $3 \times 10^7$ cycles/sample & $P_{S,i}$ & $10$ W \\
$f_{C,j}$ & $3 \times 10^8$ Hz & $m_{C,j}$ & $3 \times 10^7$ cycles/sample & $P_{C,j}$ & $0.3$ W \\
$T$ & $360$ sec & $\xi$ & $2$ & $N_0$ & $3.98 \times 10^{-21}$ W/Hz \\
$\kappa$ & $10^{-28}$ Joules$\cdot$seconds$^2$/cycles$^3$ & $B_i$ & $10$ MHz & $\gamma_{S,i},\gamma_{C,j}$ & $0.05$ \\
\hline
\end{tabular}
\vspace{-10pt}
\end{table*}

\section{Experimental Results}\label{sec:sec5}

\subsection{Satellite Megaconstellation Simulation Setup}

We simulate LEO satellite megaconstellations with technical specifications from Starlink's official release of FCC approval \cite{starlink}. We employ the 550 km shell, which has 72 orbit planes times 22 equally spaced satellites, each at an altitude of 550km and an inclination of $53^\circ$. Using the MATLAB \texttt{walkerStar} and \texttt{orbitalElements} built-in functions, we obtain the Starlink megaconstellations and orbital parameters for individual satellites. Other parameters have been proposed in a recent research \cite{gts} as summarized in Table \ref{tab:simulationsetting}. However, we need to recompute the ISL and GSL rates using the satellite megaconstellation parameters of Starlink \cite{starlink}.

\subsection{Hierarchical Federated Learning Implementation}

\textit{1) Datasets}. We experiment with the FashionMNIST dataset under both i.i.d. and non-i.i.d. cluster data distributions, resp. For i.i.d., the training samples are uniformly distributed among all clusters. For non-i.i.d., the training samples follow the Dirichlet distribution Dir($2.0$) as class priors.

\textit{2) Models}. The Crammer-Singer SVM \cite{crammersinger} is a multi-class extension of the standard binary SVM. Using $\ell_2$ regularization $(1/2)\|\mathbf{W}\|_F^2$, the objective is strongly convex with modulus 1. The ResNet model series \cite{resnet18} were designed to address the degradation problem in training DNNs, which introduce skip connections that bypass intermediate layers to effectively learn the residue. We use the lightweight ResNet-18 (having about 11.7M parameters), which has a nonconvex objective function.

\textit{3) Baselines}. To highlight the efficiency of Gauss-Seidel-type updates, we strive to make a fair comparison between our model and those that differ from ours only in that they use Jacobian-type updates. In particular, we consider the traditional HFL model (denoted as HFedAvg) \cite{ton25} and compare our algorithm with theirs in terms of communication sizes and time to train to achieve a predefined performance threshold.

\subsection{Performance Evaluation of HAFL}

\begin{figure}[htbp]
    \includegraphics[width=0.24\textwidth]{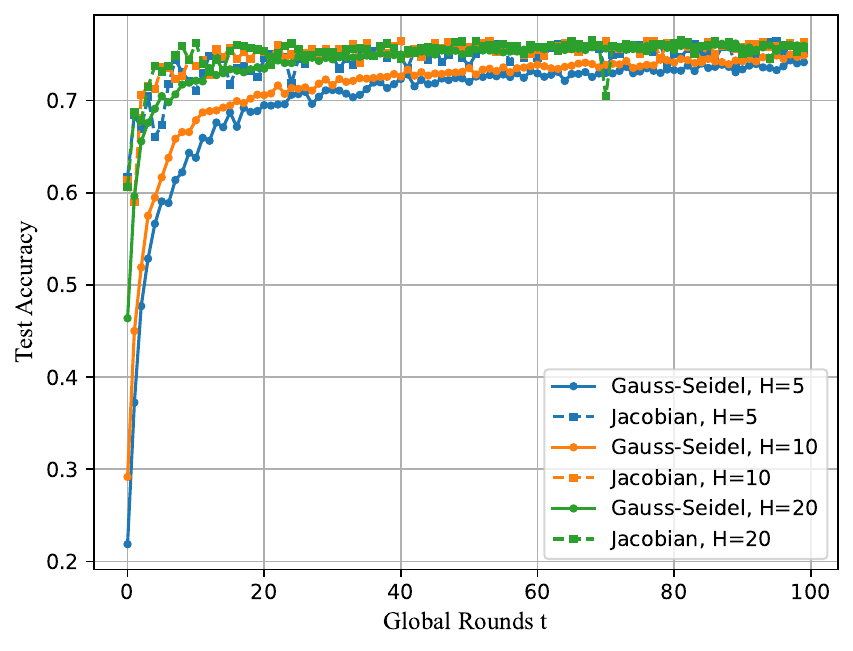}
    \includegraphics[width=0.24\textwidth]{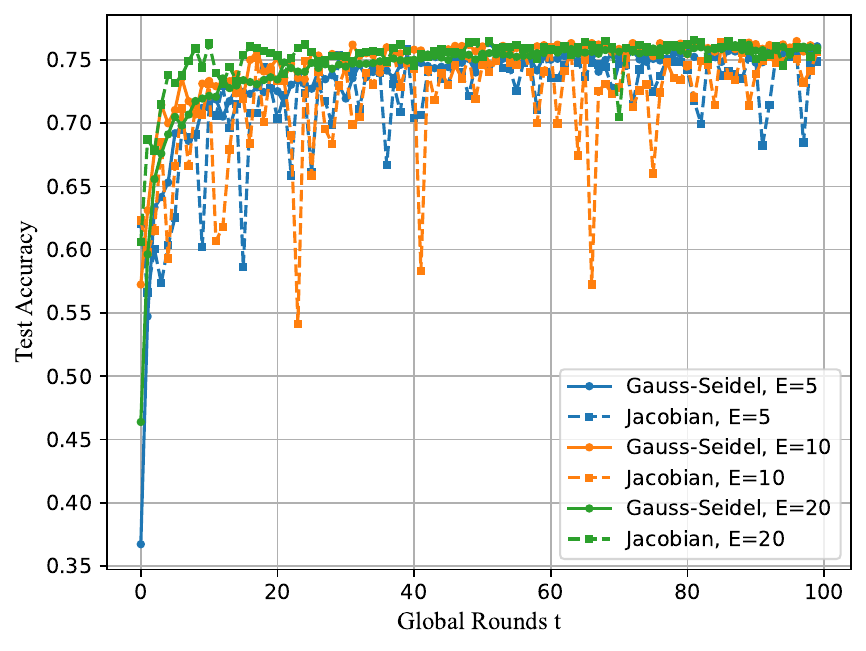}
    \vspace{10pt}
    \includegraphics[width=0.24\textwidth]{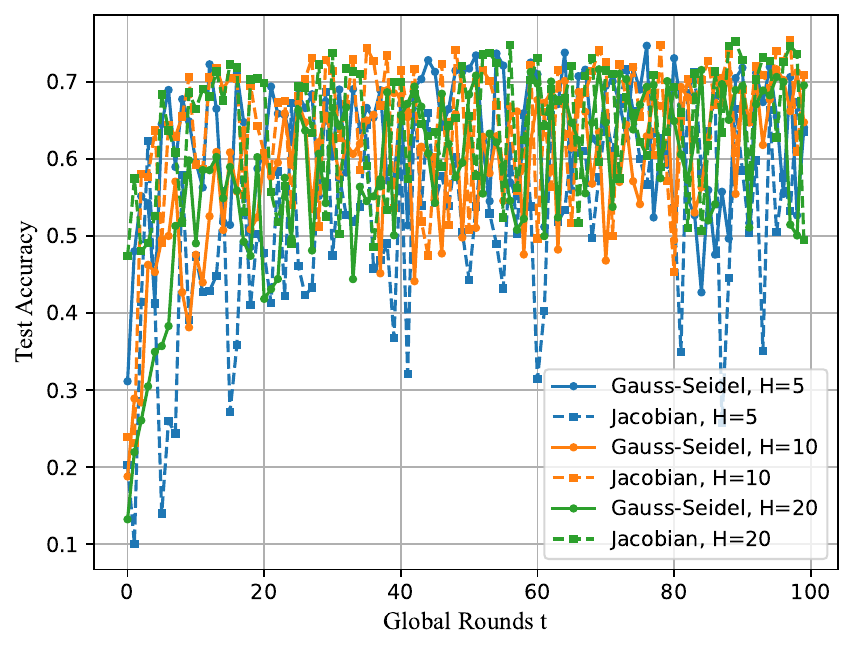}
    \includegraphics[width=0.24\textwidth]{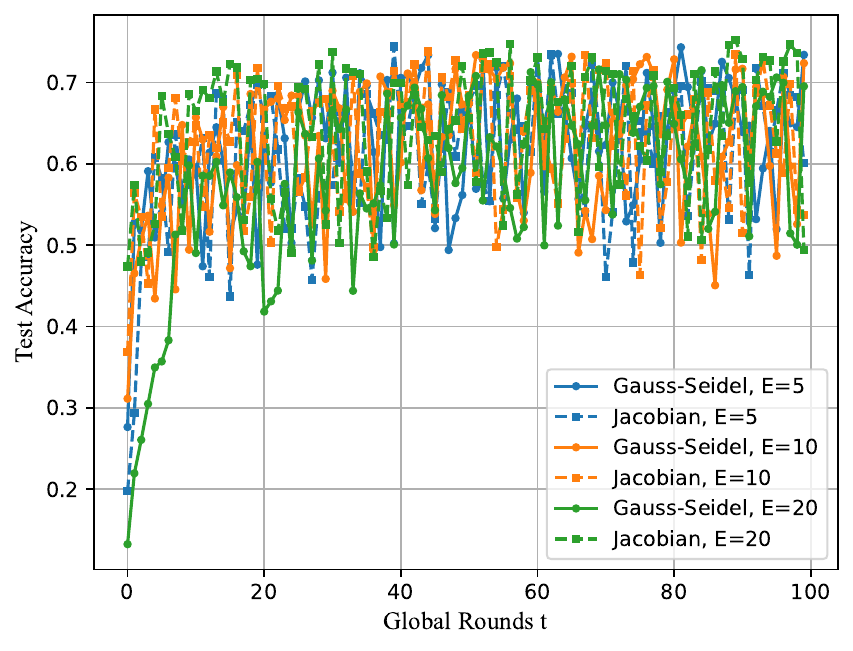}
    \caption{Jacobian and Gauss-Seidel updates for Crammer-Singer SVM with $T=50$ under i.i.d. and non-i.i.d., respectively. Upper left: i.i.d. and $E=20$, upper right: i.i.d. and $H=20$, lower left: non-i.i.d. and $E=20$, lower right: non-i.i.d. and $H=20$.}
    \label{fig:smsvm}
    \vspace{-10pt}
\end{figure}

Fig. \ref{fig:smsvm} compares the performance of Crammer-Singer SVM with HAFL and HFedAvg under i.i.d. and non-i.i.d. cluster data distributions, respectively. For moderate $E=20$ under i.i.d., performance curves become close as $H$ increases from $5$ to $20$, and become indistinguishable for $H=20$. For moderate $H=20$ under i.i.d., the performance curves of HAFL are superior to those of HFedAvg, and those of HFedAvg exhibit occasional downward spikes. That is, for i.i.d., HAFL performs on par with HFedAvg for moderate $H$ and exhibits more stable training curves. Under non-i.i.d., we observe similar patterns that HAFL performs on par with HFedAvg and exhibits more stable training curves. In practice, moderate choices of $H$ and $E$ are preferred to reduce communication cost, where HAFL performs better than or on par with HFedAvg.

\begin{figure}[htbp]
    \includegraphics[width=0.24\textwidth]{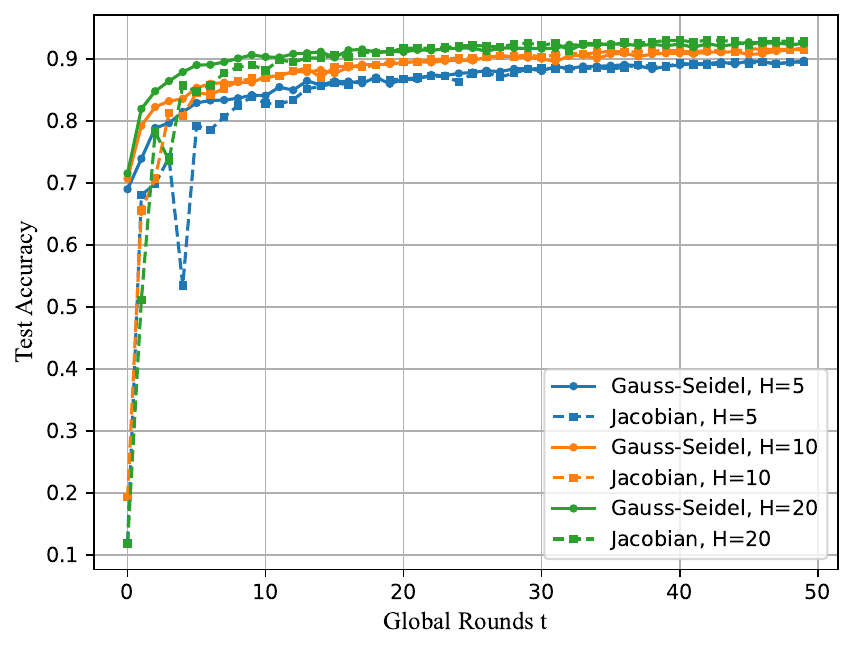}
    \includegraphics[width=0.24\textwidth]{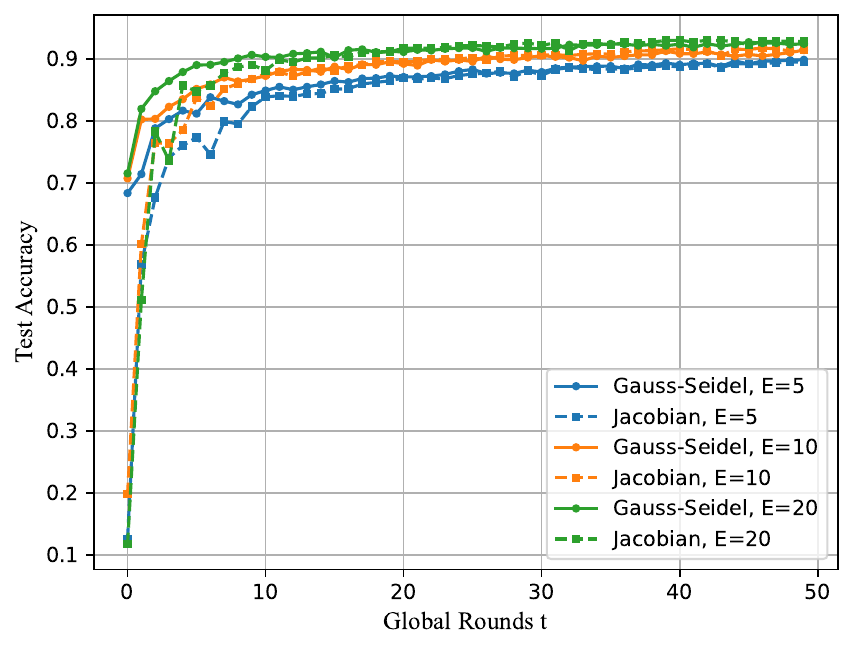}
    \vspace{10pt}
    \includegraphics[width=0.24\textwidth]{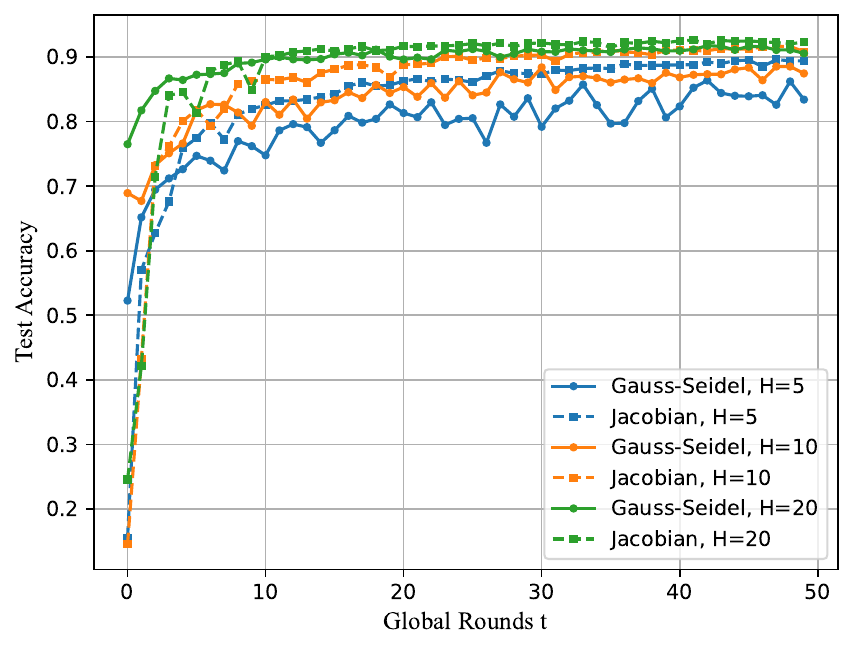}
    \includegraphics[width=0.24\textwidth]{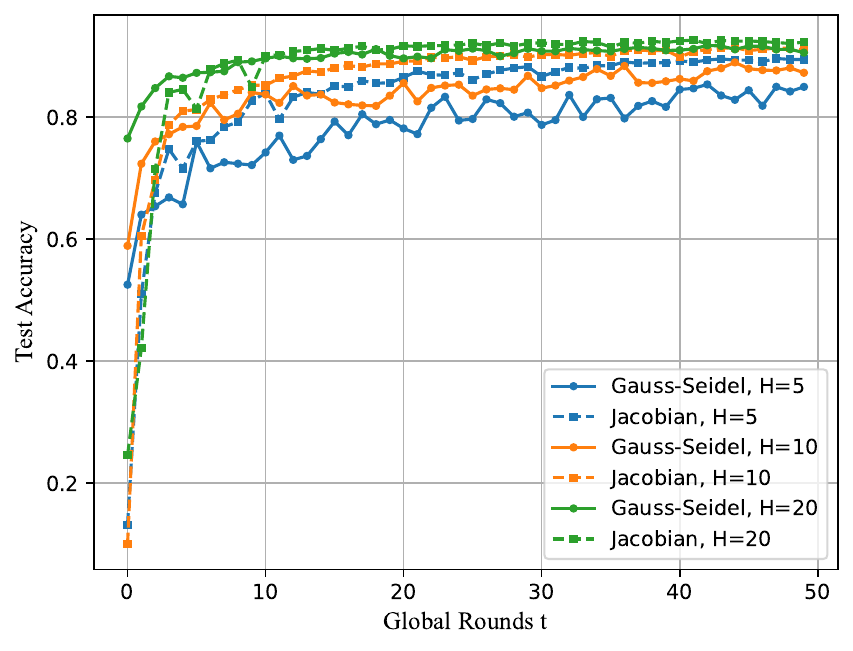}
    \caption{Jacobian and Gauss-Seidel updates for ResNet-18 with $T=50$ under i.i.d. and non-i.i.d., respectively. Upper left: i.i.d. and $E=20$, upper right: i.i.d. and $H=20$, lower left: non-i.i.d. and $E=20$, lower right: non-i.i.d. and $H=20$.}
    \label{fig:resnet18}
    \vspace{-10pt}
\end{figure}

Fig. \ref{fig:resnet18} compares the performance of ResNet-18 with HAFL and HFedAvg for under i.i.d. and non-i.i.d. cluster data distributions, respectively. Under i.i.d., the performance curves of HAFL and HFedAvg become almost indistinguishable. Under non-i.i.d., the performance curves of HAFL get close to those of HFedAvg for moderate values of $E$ or $H$. Despite such similarities, a striking benefit of HAFL is that training is fast in the first few global rounds, while HFedAvg suffers from a slow start, which implies lower latency for HAFL to achieve a predefined performance threshold and less communication cost. We highlight this aspect in Table II, where Gauss-Seidel-based HAFL significantly outperforms Jacobian-based HFedAvg with less global rounds and total per-client communication cost to achieve a predefined performance threshold.

\begin{table*}[!t]
\label{tab:haflhfedavg}
\centering
\caption{Global rounds and total per-client satellite-client communication cost (MB) for achieving testing accuracy of 70\% and 80\% on Fashion-MNIST with ResNet-18, respectively.}
\vspace{10pt}
\begin{tabular}{ccccc}
\toprule
\multicolumn{5}{c}{\textbf{ResNet-18 on Fashion-MNIST}} \\
\midrule
 & \multicolumn{2}{c}{Fully i.i.d.} & \multicolumn{2}{c}{Cell non-i.i.d., client i.i.d.} \\
 & \multicolumn{1}{c}{3 Cells} & 5 Cells & 3 Cells & 5 Cells \\
\cmidrule(lr){2-3} \cmidrule(lr){4-5}
Accuracy & HAFL / HFedAvg & HAFL / HFedAvg & HAFL / HFedAvg & HAFL / HFedAvg \\
\midrule
70\% & 1, 89.70 / 3, 234.00 & 1, 58.50 / 5, 234.00  & 1, 448.50 / 3, 234.00 & 1, 58.50 / 3, 140.40 \\
80\% & 3, 269.10 / 5, 390.00 & 2, 117.00 / 6, 280.80 & 5, 538.20 / 8, 624.00 & 3, 175.50 / 6, 280.80 \\
\bottomrule
\end{tabular}
\vspace{-10pt}
\end{table*}

\subsection{Performance Evaluation with LEO Satellites}

Our proposed satellite-assisted training with client data offloading leverages satellite-side compute and decreases cluster latency. In Section \ref{sec:sec3}, parameters $\{\alpha_j\}_{j\in\mathcal{C}_i}$ that designate the portion of local data samples of client $j$ offloaded to satellite $i$ have a significant impact on the training latency.

\begin{figure}[htbp]
    \includegraphics[width=0.24\textwidth]{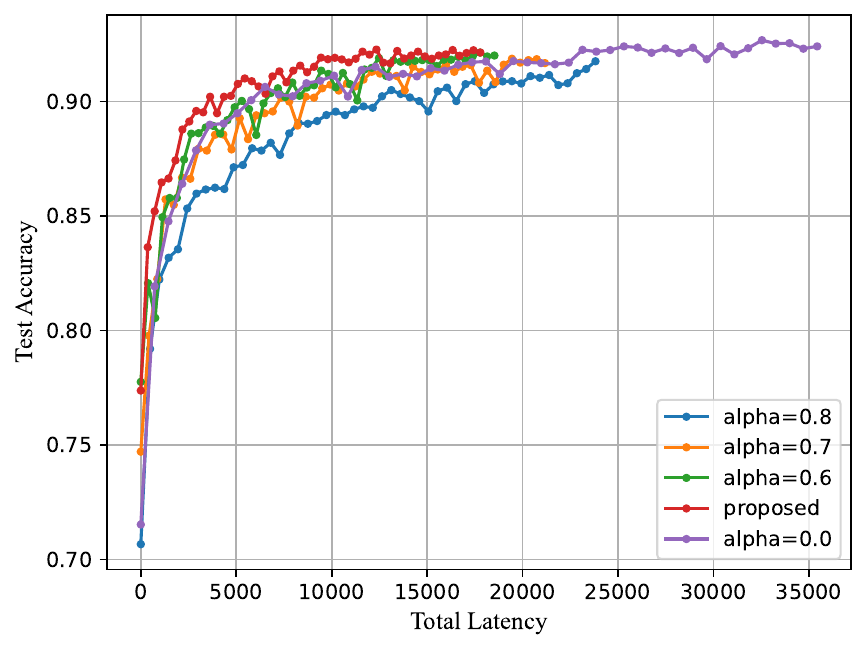}
    \includegraphics[width=0.24\textwidth]{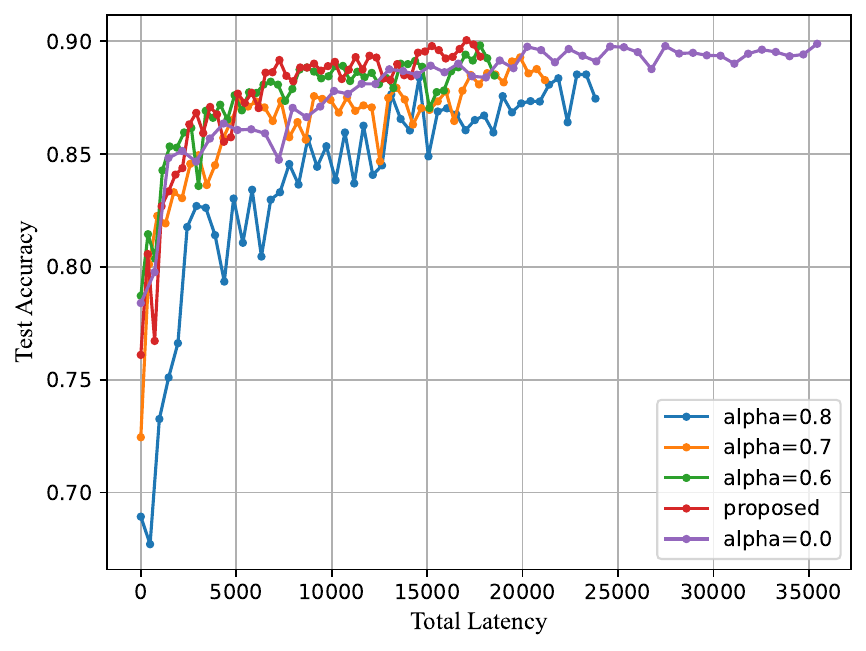}
    \vspace{5pt}
    \caption{Gauss-Seidel updates for ResNet-18 with $T=50$, $E=20$, $H=20$ under i.i.d. and non-i.i.d., respectively.}
    \label{fig:alpha}
\end{figure}
\vspace{-10pt}

Fig. \ref{fig:alpha} compares the total latency (including training and communication) in training ResNet-18 with HAFL under i.i.d. and non-i.i.d. cluster data distributions, respectively. Compared with fixed values of $\{\alpha_j\}_{j\in\mathcal{C}_i}$, solving our proposed optimization problem in Section \ref{sec:sec3} over $\{\alpha_j\}_{j\in\mathcal{C}_i}$ results in minimized total latency. Note that the final performances after $T$ global rounds, corresponding to different choices of ${\alpha_j}_{j\in\mathcal{C}_i}$, are not affected and exhibit only minor differences.

\section{Conclusion and Future Work}\label{sec:sec6}

We present HAFL, a hierarchical FL framework with asynchronous Gauss-Seidel-type updates assisted by LEO satellite megaconstellations to address the challenges of distributed learning with resource-constrained devices in remote areas. Theoretical analysis establishes convergence guarantees across various objectives, and optimization formulations minimize training latency through resource allocation and model partitioning. Experimental results validate the advantages of HAFL in achieving rapid initial convergence with competitive final accuracy, but with fewer global rounds and less communication costs. In the future, we will experiment with LLMs instead of ResNet-18 and rigorously formulate a satellite routing problem for Section \ref{sec:sec3d}.

\section*{Acknowledgments}
This research has been supported in part by the Singapore Ministry of Education Academic Research Fund Tier 2 MOE-T2EP20224-0009 and NTU startup.

\end{document}